\documentclass{article}

\usepackage[preprint]{neurips_2026}

\usepackage[utf8]{inputenc}
\usepackage[T1]{fontenc}
\usepackage{microtype}
\usepackage{graphicx}
\usepackage{subcaption}
\usepackage{booktabs}
\usepackage{hyperref}
\usepackage{url}
\usepackage{amsmath}
\usepackage{amssymb}
\usepackage{mathtools}
\usepackage{amsthm}
\usepackage[capitalize,noabbrev]{cleveref}
\usepackage{enumitem}
\usepackage{placeins}
\usepackage[table]{xcolor}
\usepackage{tcolorbox}
\usepackage{algorithm}
\usepackage{algorithmic}
\usepackage{tikz}
\usetikzlibrary{decorations.pathreplacing,arrows.meta,shapes.geometric}
\usepackage{pifont}
\usepackage{multirow}
\usepackage{array}
\usepackage{tabularx}
\usepackage{makecell}
\usepackage{longtable}

\theoremstyle{plain}

\theoremstyle{definition}

\theoremstyle{remark}

\newcommand{\cmark}{\ding{51}}
\newcommand{\xmark}{\ding{55}}

\begin{document}

\title{Projected Autoregression: Autoregressive Language Generation in Continuous State Space}

\author{%
Oshri Naparstek \\
IBM Research, Haifa, Israel \\
\texttt{oshri.naparstek@ibm.com}
}

\maketitle

%%%%%%%%%%%%%%%%%%%%%%%%%%%%%%%%
% ABSTRACT
%%%%%%%%%%%%%%%%%%%%%%%%%%%%%%%%
\begin{abstract}
Standard autoregressive language models generate text by repeatedly selecting a discrete next token, coupling prediction with irreversible commitment at every step.
We show that token selection is not the only viable autoregressive interface.

\textbf{Projected Autoregression} replaces token selection with continuous prediction in embedding space followed by discrete projection at commitment time.
The model predicts next-token vectors via regression and contrastive objectives, while discrete tokens arise only by nearest-neighbor projection.
An optional mutable suffix (``liquid tail'') enables iterative refinement before commitment, but the central change is more basic: next-step prediction is continuous, and discrete tokens are produced only as a downstream interface.

Projected Autoregression establishes a concrete alternative to token-selection autoregression: language generation can be organized around continuous-state prediction with delayed discrete commitment.
Refinement remains local to a short causal suffix within a left-to-right causal process, rather than a sequence-wide denoising process.
This separation has two consequences.
First, it induces a \emph{distinct generation regime}: even with immediate projection ($K{=}1$), continuous prediction yields text structure and dynamics that differ from tested token-space AR baselines, including a compute-matched best-of-16 reranking baseline.
Second, it exposes a \emph{continuous control surface} inside autoregressive generation: direction rate, history noise, delayed commitment, state-space guidance, and embedding geometry act directly on the evolving generative state before token commitment.
Taken together, these results place repeated token selection within a larger family of autoregressive interfaces and expose continuous state space as a broader algorithmic design space for language generation.
\end{abstract}

%%%%%%%%%%%%%%%%%%%%%%%%%%%%%%%%
% 1. INTRODUCTION
%%%%%%%%%%%%%%%%%%%%%%%%%%%%%%%%

\section{Introduction}
\label{sec:intro}

Transformer language models generate text autoregressively by selecting a token at each step from a fixed vocabulary \cite{vaswani2017attention}.
In standard decoding, the model produces token scores and immediately commits to a discrete token (via greedy choice or sampling), making the prefix a sequence of token indices that fully condition subsequent generation.
This forces uncertainty to be resolved through discrete choices at every step, with diversity and stability controlled largely by token-level heuristics.

In discrete token space, delayed commitment is intrinsically awkward: it requires non-local search procedures such as beam search, suffix rewriting, or masked re-decoding---none of which provide a simple local refinement rule compatible with left-to-right generation.
In continuous embedding space, the same operation becomes natural: predict a vector, move the state toward it, and project to a discrete token only when needed.

We propose \emph{Projected Autoregression}, a framework built around this asymmetry.
The model predicts in embedding space via regression and contrastive objectives, and maintains a short, mutable suffix---a \emph{liquid tail}---whose vectors are iteratively refined before discrete commitment via nearest-neighbor projection.
Importantly, even immediate projection ($K{=}1$) from continuous predictions already induces a distinct generation regime; the liquid tail ($K{>}1$) further improves diversity and suppresses degeneration by sustaining local refinement before commitment.

This yields a generation process that remains fully autoregressive and causal, yet differs from both standard decoding and diffusion-style generation:
uncertainty is represented geometrically and can persist across refinement steps before commitment, while refinement is local to a short suffix rather than global denoising.
Projected Autoregression is a continuous-state autoregressive interface with delayed discrete commitment.
Generation remains strictly causal and left-to-right: the backbone is a standard causal Transformer, and only a short future suffix is refined before commitment.
The result reframes standard token selection as one point in a broader space of continuous-state autoregressive algorithms.
Once prediction happens in continuous space and commitment is deferred, operations that are awkward or unavailable in token space become natural local transformations of the evolving generative state.

\begin{figure*}[t]
\centering
\includegraphics[width=0.92\textwidth]{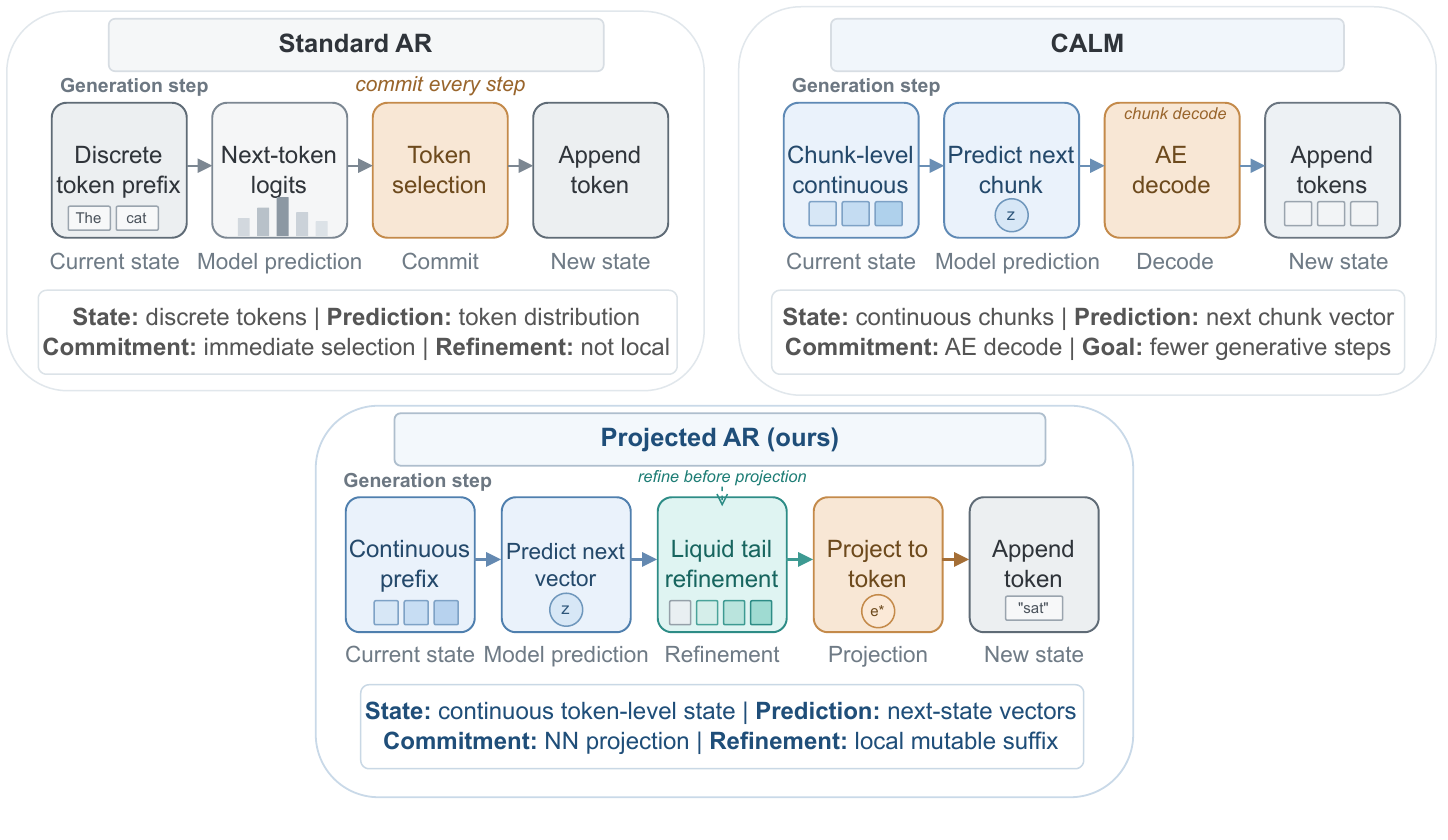}
\caption{\textbf{Three autoregressive interfaces.}
Standard AR operates entirely in discrete space.
CALM predicts continuous chunk vectors decoded via autoencoder.
Projected AR predicts continuous token-level vectors with optional local refinement (liquid tail), committed via nearest-neighbor projection.
}
\label{fig:interfaces}
\end{figure*}

Importantly, commitment is not tied to probability concentration: neighborhood narrowing (Sec.~\ref{sec:ablation}) shows that a local causal suffix converges geometrically to a small set of plausible tokens well before discrete commitment.

\textbf{Contributions.}
\begin{itemize}
  \item We introduce \emph{Projected Autoregression}, a continuous-state autoregressive interface that replaces token selection with continuous prediction in embedding space followed by discrete projection, separating prediction from commitment.
  \item We demonstrate that separating prediction from commitment exposes a family of \emph{continuous-state control knobs}---direction rate, history noise, delayed commitment, state-space guidance, and embedding geometry---that operate directly on the evolving generative state.
  \item We show that this interface produces a measurably distinct generation regime. A $K{=}1$ decomposition localizes the effect to continuous prediction itself, while a compute-matched AR baseline highlights the resulting differences in text.
  \item We provide a mechanistic analysis revealing progressive neighborhood narrowing in the liquid tail, showing how continuous-state refinement approaches discrete commitment while retaining a set of plausible token continuations.
\end{itemize}

%%%%%%%%%%%%%%%%%%%%%%%%%%%%%%%%
% 2. RELATED WORK
%%%%%%%%%%%%%%%%%%%%%%%%%%%%%%%%
\section{Related Work}

Prior work has explored continuous states in language modeling, but in substantially different roles.
Coconut~\cite{hao2024training} uses continuous latent states as an auxiliary reasoning substrate: hidden states are fed back as ``continuous thoughts'' to support planning and backtracking, rather than serving as the primary token-generation interface.
TiDAR~\cite{liu2025tidar} combines diffusion-based parallel drafting with autoregressive final output generation, positioning continuous states within a hybrid sequence-level drafting architecture aimed at throughput.
More recently, Continuous Autoregressive Language Models (CALM)~\cite{shao2025calm} replace discrete next-token prediction with continuous next-vector prediction at the level of compressed token chunks, decoded via a learned autoencoder; their focus is improving the performance--compute trade-off by reducing the number of generative steps.
Diffusion language models~\cite{dream2025,diffucoder2025,sahoo2024simple,lou2023discrete} operate via global denoising over the full sequence, enabling parallel generation but precluding streaming.
Another nearby line is draft-and-verify acceleration, including speculative decoding and related lookahead methods~\cite{leviathan2023fast}.
These methods maintain \emph{discrete} draft hypotheses that are later accepted or rejected by a standard token model, so the underlying autoregressive interface remains token selection.
They accelerate commitment in token space, whereas Projected Autoregression refines a mutable continuous pre-commitment state before projection.

Projected Autoregression shares with CALM the move from discrete to continuous prediction, but differs in granularity (individual tokens, not chunks), decoding mechanism (nearest-neighbor projection, not autoencoder), and research question.
Where CALM asks whether continuous AR can be more efficient, we ask what continuous token-level prediction \emph{does} to the generation regime---showing that even immediate projection ($K{=}1$) induces measurably different text structure, and characterizing the mechanistic and geometric consequences of this interface change.
Relative to diffusion LLMs, the central contrast is \emph{local causal refinement} versus \emph{global sequence denoising}: only a short suffix remains mutable, and committed prefix tokens are never revisited.

%%%%%%%%%%%%%%%%%%%%%%%%%%%%%%%%
% 3. METHOD
%%%%%%%%%%%%%%%%%%%%%%%%%%%%%%%%
%%%%%%%%%%%%%%%%%%%%%%%%%%%%%%%%
% METHOD
%%%%%%%%%%%%%%%%%%%%%%%%%%%%%%%%
\section{Projected Autoregression}
\label{sec:method}

We introduce a continuous-variable formulation of autoregressive language modeling in which tokens are represented and generated as vectors in embedding space, and discrete commitment is deferred through iterative refinement of a short mutable suffix.
This section formalizes the representation, generation dynamics, and training objective underlying the proposed framework.

\subsection{Continuous Token Representation}
Let $\mathcal{V}$ denote a discrete vocabulary of size $|\mathcal{V}|$, and let $E \in \mathbb{R}^{|\mathcal{V}| \times d}$ be a fixed embedding matrix, where each row $e_i \in \mathbb{R}^d$ corresponds to a token embedding.
We assume embeddings are $\ell_2$-normalized and scaled to a fixed radius $R$.

Rather than predicting a categorical distribution over $\mathcal{V}$, the model predicts continuous vectors $z_t \in \mathbb{R}^d$ at each position $t$.
Discrete tokens are recovered only at commitment time by projecting continuous vectors onto the vocabulary embedding set.

\subsection{Autoregressive Vector Prediction}
Given a sequence of previously committed token vectors $\{z_1, \dots, z_{t-1}\}$, the model predicts a continuous vector $\hat{z}_t$ via an autoregressive function
\begin{equation}
\hat{z}_t = f_\theta(z_1, \dots, z_{t-1}),
\end{equation}
where $f_\theta$ is implemented as a causal Transformer operating directly in embedding space.
Importantly, $\hat{z}_t$ is not immediately discretized and may evolve over time before commitment.

\subsection{Conditioning on Maturation State}
\label{sec:conditioning}

To enable the model to behave appropriately at different stages of the refinement process, we condition on two quantities: a \emph{maturity coordinate} $\alpha_t \in [0,1]$ at each position, and the tail length $K$.

\textbf{Maturity coordinate $\alpha$.}
We intentionally use $\alpha$ as a single scalar that jointly parameterizes (i) the token's position within the liquid tail and (ii) its uncertainty.
Intuitively, newly introduced tail tokens are maximally liquid ($\alpha \approx 0$), while tokens approaching commitment are nearly stabilized ($\alpha \approx 1$).
During training, uncertainty is realized through an $\alpha$-dependent corruption magnitude $\sigma(\alpha)$ (defined in Sec.~\ref{sec:training_obj}).
During inference, the same $\alpha$ also determines the refinement step size $\eta(\alpha)$ (defined in Sec.~\ref{sec:maturation}).
This coupling between position and uncertainty is the core inductive bias of Projected Autoregression.

\textbf{Maturity conditioning.}
The model is conditioned on $\alpha$ and $K$ so it can distinguish committed context from maturing predictions (architecture illustrated in Appendix~\ref{sec:app_arch}).
We embed $\alpha_t$ using a sinusoidal encoding (as commonly used in diffusion models) followed by a learned MLP, and add the result to the token representation:
\begin{equation}
h_t \leftarrow h_t + \mathrm{MLP}(\mathrm{SinEmb}(\alpha_t)).
\end{equation}
This allows the model to distinguish between stabilized (prefix-like) tokens ($\alpha \approx 1$) and liquid tail tokens ($\alpha \approx 0$).

\textbf{Tail-length conditioning.}
We further condition the model on the current tail length $K$ via feature-wise linear modulation (FiLM).
A learned embedding of $K$ is projected to produce scale and shift parameters $(\gamma, \beta)$, which modulate the hidden representations:
\begin{equation}
h \leftarrow (1 + \gamma) \odot h + \beta.
\end{equation}
This global conditioning allows the model to adjust its predictions based on how much context is committed versus liquid.

\subsection{Projected Autoregression}
\label{sec:maturation}

To decouple prediction from commitment, we maintain a \emph{refinement buffer} of length $K$, referred to as the \emph{liquid tail}.
At any generation step, the model maintains a sequence
\[
(z_1, \dots, z_{t-K}, \tilde{z}_{t-K+1}, \dots, \tilde{z}_t),
\]
where the final $K$ vectors are uncommitted and continuously updated.

\textbf{Tail maturity schedule.}
For a tail of length $K$ at time $t$, we assign each tail position $i\in\{t-K+1,\dots,t\}$ a maturity value
\begin{equation}
\alpha_i = \frac{t-i}{\max(1,\,K-1)},
\end{equation}
so the newest token has $\alpha \approx 0$ and the oldest tail token has $\alpha \approx 1$.
Other monotone schedules are possible; we use this simple linear coupling throughout.

\textbf{Iterative refinement.}
At each step, the backbone predicts $\hat{z}_i = f_\theta(\cdot)$ for positions in the tail, and tail vectors are refined by a contraction update
\begin{equation}
\tilde{z}_i \leftarrow \tilde{z}_i + \eta(\alpha_i)\,(\hat{z}_i - \tilde{z}_i),
\end{equation}
where $\eta(\alpha)\in(0,1]$ is an $\alpha$-dependent refinement step size derived from the same maturity coordinate.
We use $\eta(\alpha)$ increasing with $\alpha$, so older tail positions (larger $\alpha$) stabilize more aggressively while newly introduced vectors evolve slowly.
A simple choice is
\begin{equation}
\eta(\alpha) = \eta_{\min} + (\eta_{\max}-\eta_{\min})\,\alpha^p,
\end{equation}
with $\eta_{\min}>0$, $\eta_{\max}\le 1$, and $p\ge 1$.

This process allows uncertainty to be expressed geometrically as distance in embedding space and resolved incrementally rather than through instantaneous sampling.

\subsection{Discrete Commitment via Projection}
Once a token vector reaches the front of the liquid tail, it is committed by projection onto the embedding matrix:
\begin{equation}
x_t = \arg\max_{i \in \mathcal{V}} \langle z_t, e_i \rangle.
\end{equation}
The committed vector is then replaced by its corresponding embedding $e_{x_t}$ and becomes part of the fixed autoregressive context.
Although commitment uses an argmax operation, stochasticity arises implicitly through the continuous refinement dynamics rather than explicit sampling.

\subsection{Training Objective}
\label{sec:training_obj}

A pure regression objective on continuous vectors leads to mode averaging and collapse.
To stabilize training and align continuous predictions with discrete token identity, we combine a mean-squared error objective with a contrastive loss.

\textbf{Simulated liquid-tail corruption.}
During training, we simulate tail uncertainty by corrupting the ground-truth embedding with an $\alpha$-dependent noise magnitude:
\begin{equation}
\tilde{z}_t = e_{x_t} + \sigma(\alpha_t)\,\epsilon,
\qquad \epsilon \sim \mathcal{N}(0,I),
\end{equation}
where $\sigma(\alpha)$ is a monotone schedule (we use $\sigma(\alpha)=\sigma_{\max}(1-\alpha)$ unless stated otherwise).
The model observes the corrupted tail vector $\tilde{z}_t$ together with its maturity value $\alpha_t$ and tail length $K$, and predicts $\hat{z}_t$.

Given a predicted vector $\hat{z}_t$ and its ground-truth embedding $e_{x_t}$, we minimize
\begin{equation}
\mathcal{L}_{\text{reg}} = \|\hat{z}_t - e_{x_t}\|_2^2,
\end{equation}
alongside a contrastive InfoNCE loss
\begin{equation}
\mathcal{L}_{\text{NCE}} = -\log
\frac{\exp(\langle \hat{z}_t, e_{x_t} \rangle / \tau)}
{\sum_{j \in \mathcal{N}} \exp(\langle \hat{z}_t, e_j \rangle / \tau)},
\end{equation}
where $\mathcal{N}$ is a set of negative samples and $\tau$ is a temperature parameter.

The final training objective is
\begin{equation}
\mathcal{L} = \mathcal{L}_{\text{reg}} + \lambda \mathcal{L}_{\text{NCE}}.
\end{equation}

This contrastive component prevents collapse toward frequent tokens and anchors continuous predictions to discrete semantic identities without requiring a softmax likelihood.
\textbf{Computational overhead.}
\label{sec:compute}
The liquid tail introduces a $K$-fold overhead per token (recomputing $K$ positions per step while reusing the prefix KV cache).
Since $K{=}1$ already produces most structural benefits (Sec.~\ref{sec:ablation}), this overhead is optional: practitioners can use $K{=}1$ for no overhead, or $K{>}1$ when extra diversity is desired.

%%%%%%%%%%%%%%%%%%%%%%%%%%%%%%%%
% 4. TRAINING OBJECTIVE
%%%%%%%%%%%%%%%%%%%%%%%%%%%%%%%%
%%%%%%%%%%%%%%%%%%%%%%%%%%%%%%%%
% TRAINING AND GENERATION
%%%%%%%%%%%%%%%%%%%%%%%%%%%%%%%%
\section{Training and Generation}
\label{sec:training}

This section describes how the proposed model is trained and how autoregressive generation is performed at inference time.
Although training and generation operate under different constraints, both are governed by the same underlying liquid-tail refinement dynamics.

\subsection{Training with Simulated Maturation}
During training, the model is exposed to partially matured token representations to encourage robustness to uncertainty and to align training dynamics with inference-time behavior.
Given a ground-truth token sequence $(x_1, \dots, x_T)$, we first map tokens to their embedding representations $(e_{x_1}, \dots, e_{x_T})$.

To simulate the presence of a liquid tail, we perturb a suffix of length $K$ by mixing the ground-truth embeddings with isotropic noise in embedding space.
Earlier tokens remain fixed, while later tokens are progressively corrupted, mimicking different stages of refinement.
This procedure exposes the model to inputs ranging from fully committed tokens to highly uncertain representations.

The model is trained to predict the next-step continuous vector $\hat{z}_{t+1}$ given the current sequence of committed and uncommitted vectors, using the combined regression and contrastive objective described in Section~\ref{sec:method}.
\textbf{Loss weighting.}
To prevent the model from overweighting highly corrupted positions where the target is inherently ambiguous, we weight the loss at each position by $(1 - \alpha_t)$, where $\alpha_t$ is the noise level.
Positions with low noise (near commitment) contribute more to the gradient, while highly uncertain positions contribute less.
\subsection{Noise Injection and Stability}
Noise injection during training serves two complementary purposes.
First, it regularizes the model by preventing over-reliance on exact embedding vectors.
Second, it approximates the distribution of uncommitted token states encountered during generation.

Importantly, noise is bounded and scaled such that vector norms evolve gradually over time.
This ensures that uncertainty is resolved through gradual refinement rather than abrupt stochastic jumps, and avoids the training--inference mismatch commonly encountered when noise is injected only at sampling time.

\subsection{Autoregressive Generation}
At inference time, generation proceeds autoregressively from left to right.
Given an initial prompt, the corresponding token embeddings are inserted into the sequence as committed vectors.
A liquid tail of length $K$ is initialized with low-norm random vectors, representing highly uncertain token states.

At each generation step, the model predicts updated continuous vectors for the entire sequence.
Vectors in the liquid tail are updated according to the refinement rule, while committed tokens remain fixed.
Once a vector reaches the front of the liquid tail, it is discretized via projection onto the vocabulary embedding matrix and committed permanently.

This process yields a stream of discrete tokens, while internally maintaining a continuous representation that evolves over time.
\textbf{Classifier-free guidance.}
At inference time, we optionally apply guidance to sharpen predictions toward the committed (conditioned) context.
We compute two forward passes that differ only in the attention mask:
(i) a \emph{context-aware} pass with the standard full causal mask, and
(ii) a \emph{context-masked} pass in which tail positions are prevented from attending to the committed prefix (i.e., tail-only attention).
We then combine the resulting predictions for each position $t$:
\begin{equation}
\hat{z}_t = \hat{z}_t^{\text{cm}} + s \cdot \big(\hat{z}_t^{\text{ca}} - \hat{z}_t^{\text{cm}}\big),
\end{equation}
where $s \geq 1$ is the guidance scale, $\hat{z}_t^{\text{ca}}$ denotes the context-aware prediction,
and $\hat{z}_t^{\text{cm}}$ denotes the context-masked prediction.
This encourages the evolving tail to remain consistent with the committed history.

\subsection{Generation Algorithm}
Algorithm~\ref{alg:generation} summarizes the generation process.

\begin{algorithm}[t]
\caption{Generation with Projected Autoregression}
\label{alg:generation}
\begin{algorithmic}[1]
\REQUIRE Prompt embeddings $(e_{x_1}, \dots, e_{x_n})$, tail length $K$, guidance scale $s$
\STATE Initialize committed sequence $\mathbf{z}_{1:n} \leftarrow (e_{x_1}, \dots, e_{x_n})$
\STATE Initialize liquid tail $\tilde{\mathbf{z}}_{n+1:n+K}$ with random low-norm vectors
\STATE Construct maturity profile $\boldsymbol{\alpha}$ with $\alpha_t=1$ for $t \le n$ and $\alpha_t$ fading over the tail
\WHILE{not end-of-sequence}
    \STATE $\hat{\mathbf{z}}^{\text{ca}} \leftarrow f_\theta(\mathbf{z}, \boldsymbol{\alpha}, K)$ \COMMENT{context-aware: full causal mask}
    \STATE $\hat{\mathbf{z}}^{\text{cm}} \leftarrow f_\theta(\mathbf{z}, \boldsymbol{\alpha}, K)$ \COMMENT{context-masked: tail cannot attend to prefix}
    \STATE $\hat{\mathbf{z}} \leftarrow \hat{\mathbf{z}}^{\text{cm}} + s \cdot (\hat{\mathbf{z}}^{\text{ca}} - \hat{\mathbf{z}}^{\text{cm}})$ \COMMENT{guidance}
    \STATE Update tail: $\tilde{z}_i \leftarrow \tilde{z}_i + \eta(\alpha_i)\,(\hat{z}_i - \tilde{z}_i)$ for $i \in \{n{+}1,\dots,n{+}K\}$
    \STATE Commit front token: $x_{n+1} \leftarrow \arg\max_j \langle \tilde{z}_{n+1}, e_j \rangle$
    \STATE Replace: $z_{n+1} \leftarrow e_{x_{n+1}}$; append a new embryo to keep tail length $K$
    \STATE $n \leftarrow n + 1$; update $\boldsymbol{\alpha}$
\ENDWHILE
\STATE \textbf{return} Generated token sequence $(x_1, \dots, x_n)$
\end{algorithmic}
\end{algorithm}

%%%%%%%%%%%%%%%%%%%%%%%%%%%%%%%%
% 5. EXPERIMENTS
%%%%%%%%%%%%%%%%%%%%%%%%%%%%%%%%
\section{Experiments}
\label{sec:experiments}

We evaluate Projected Autoregression on a modern 3B-parameter backbone against matched autoregressive baselines.
We address two questions.
\textbf{First}, does Projected Autoregression instantiate a genuinely different autoregressive interface with its own operating regime?
\textbf{Second}, does separating prediction from commitment expose practically meaningful control knobs inside autoregressive generation?

\subsection{Experimental Setup}
\label{sec:exp_setup}

\textbf{Model and baselines.}
We apply Projected Autoregression via LoRA~\cite{hu2022lora} on Granite 3B Code Base~\cite{mishra2024granite}, trained on FineWeb~\cite{penedo2024fineweb} with the MSE~+~InfoNCE objective (Sec.~\ref{sec:training_obj}).
As baseline, we train a standard AR LoRA adapter on the \emph{same} backbone with \emph{identical} configuration (rank~8, $\alpha{=}32$) and data, isolating the effect of the training objective.
We evaluate AR under greedy~(B1), greedy+repetition penalty~(B2), top-$p{=}0.95$~(B4), and a compute-matched best-of-16 (16 candidates, select lowest perplexity---same forward-pass budget as $K{=}16$).
Projected Autoregression uses deterministic argmax with $K{=}16$; 50 prompts, 150 tokens/steps per response.

\subsection{Main Results}
\label{sec:main_results}

\begin{table}[t]
\centering
\small
\setlength{\tabcolsep}{3pt}
\renewcommand{\arraystretch}{1.15}
\begin{tabular}{llccccc}
\toprule
& \textbf{Method} & \textbf{Det.} & \textbf{srep4}$\downarrow$ & \textbf{d-1}$\uparrow$ & \textbf{MAUVE}$\uparrow$ & \textbf{info-d}$\uparrow$ \\
\midrule
M1 & \textbf{Proj.\ AR} $K{=}16$ & \cmark & \textbf{.001} & \textbf{.911} & .849 & \textbf{12.0} \\
\midrule
B1 & AR greedy & \cmark & .480 & .314 & .218 & 6.9 \\
B2 & AR greedy+rep & \cmark & .000 & .873 & \textbf{.953} & 10.1 \\
B4 & AR top-$p$=0.95 & \xmark & .006 & .693 & .946 & 9.2 \\
& AR best-of-16 & \xmark & .058 & .624 & --- & 9.0 \\
\bottomrule
\end{tabular}
\caption{\textbf{Main results.}
All models share the same Granite~3B backbone and FineWeb training data.
Projected Autoregression achieves near-zero repetition, highest diversity, and highest information density under deterministic argmax, while AR heuristics achieve higher MAUVE. AR best-of-16 (compute-matched) does not reproduce the same operating regime.
}
\label{tab:main}
\end{table}

Table~\ref{tab:main} shows that Projected Autoregression occupies a different operating regime from AR, leading on repetition, diversity, and information density while AR heuristics lead on MAUVE; compute-matched AR best-of-16 does not reproduce the shift.

\subsection{Decomposition: Interface vs.\ Refinement}
\label{sec:ablation}

A natural question is whether the regime shift stems from the continuous prediction interface itself or from the multi-step refinement that the liquid tail provides.
To disentangle these, we compare $K{=}1$ (immediate projection---same model, no refinement) against $K{=}16$ and the AR baselines.

\begin{table}[t]
\centering
\small
\setlength{\tabcolsep}{3pt}
\renewcommand{\arraystretch}{1.15}
\begin{tabular}{lcccccc}
\toprule
\textbf{Config} & \textbf{srep4}$\downarrow$ & \textbf{d-1}$\uparrow$ & \textbf{info-d}$\uparrow$ & \textbf{disc.}$\uparrow$ & \textbf{conn.}$\uparrow$ & \textbf{qw}$\uparrow$ \\
\midrule
$K{=}1$ (immediate) & .002 & .888 & 13.8 & 0.68 & 2.02 & 2.67 \\
$K{=}16$ (delayed) & \textbf{.001} & \textbf{.911} & 12.0 & 0.54 & 1.84 & 2.57 \\
\midrule
AR best-of-16 & .058 & .624 & 9.0 & 0.12 & 0.53 & 0.83 \\
AR greedy+rep & .000 & .873 & 10.1 & 0.28 & 1.72 & 1.91 \\
\bottomrule
\end{tabular}
\caption{\textbf{Decomposition: interface change vs.\ delayed commitment.}
$K{=}1$ uses the same continuous-space model but commits immediately after one update step.
$K{=}1$ already produces a distinct operating point.
$K{=}16$ adds diversity and anti-degeneration.
\textbf{disc.}~=~discourse markers/100w; \textbf{conn.}~=~connectives/100w; \textbf{qw}~=~question words/100w.}
\label{tab:ablation}
\end{table}

Table~\ref{tab:ablation} shows that the regime shift already appears at $K{=}1$, before multi-step tail refinement.
$K{=}16$ adds diversity (d-1: .911 vs.\ .888) and further reduces repetition.
The broader significance is that \emph{the continuous prediction interface opens a design space} in which $K$ and other parameters become controllable knobs (Sec.~\ref{sec:control_surface}).

\subsection{Mechanistic Analysis: Neighborhood Narrowing}
\label{sec:mechanistic}

To understand \emph{why} delayed commitment helps, we analyze the dynamics of the liquid tail.
We track two probes across maturation steps:
\textbf{SpreadNet} (cosine distance from each tail vector to the centroid of its top-$k$ nearest vocabulary embeddings) and
\textbf{CentroidNet} (cosine similarity between the tail vector and that centroid).

\begin{table}[t]
\centering
\small
\setlength{\tabcolsep}{5pt}
\renewcommand{\arraystretch}{1.15}
\begin{tabular}{lcc}
\toprule
\textbf{$K$} & \textbf{$\Delta$SpreadNet}$\downarrow$ & \textbf{$\Delta$CentroidNet}$\uparrow$ \\
\midrule
2  & $-0.037$ & $+0.080$ \\
16 & $-0.088$ & $+0.181$ \\
\bottomrule
\end{tabular}
\caption{\textbf{Neighborhood narrowing strengthens up to $K{\approx}16$.}
Mean probe change from first to final refinement step; larger-magnitude shifts indicate stronger convergence toward a compact neighborhood of plausible tokens.}
\label{tab:mechanistic}
\end{table}

SpreadNet decreases monotonically: the neighborhood of candidate tokens \emph{shrinks} during maturation.
CentroidNet increases: the tail converges toward the neighborhood centroid.
Both effects strengthen substantially from $K{=}2$ to $K{=}16$, after which the appendix $K$-sweep indicates saturation.
This reveals the core mechanism: the liquid tail converges toward the \emph{centroid of a narrowing cloud of plausible tokens}, not toward any single token embedding.
Continuous-state generation therefore explores local semantic neighborhoods before commitment.

\subsection{Embedding Geometry Controls Register Stability}
\label{sec:register}

With frozen (pretrained) embeddings, generated text exhibits \emph{register drift}: text shifts mid-generation from expository to promotional or organizational voice (e.g., ``This is where our team comes in!'').
To test whether this is inherent to continuous prediction, we compare frozen vs.\ learned embeddings (where the embedding space is trained alongside the continuous-prediction adapter) across both $K{=}1$ and $K{=}16$.

\begin{table}[t]
\centering
\small
\setlength{\tabcolsep}{4pt}
\renewcommand{\arraystretch}{1.15}
\begin{tabular}{lcccc}
\toprule
& \multicolumn{2}{c}{\textbf{Frozen emb.}} & \multicolumn{2}{c}{\textbf{Learned emb.}} \\
\cmidrule(lr){2-3} \cmidrule(lr){4-5}
& $K{=}1$ & $K{=}16$ & $K{=}1$ & $K{=}16$ \\
\midrule
Register-consistent & 4/10 & 2/10 & \textbf{8/10} & \textbf{8/10} \\
Register drift & 3/10 & 5/10 & 0/10 & 0/10 \\
Fabricated attribution & 3/10 & 3/10 & 2/10 & 2/10 \\
\bottomrule
\end{tabular}
\caption{\textbf{Register stability vs.\ embedding geometry.}
In this manual annotation (10 prompts per condition), learned embeddings substantially reduce register drift at both $K{=}1$ and $K{=}16$.
Since the effect persists at $K{=}1$, register drift depends on the geometry of the token space, not on delayed commitment.
Learned embeddings produce slightly less discourse-rich text (Table~\ref{tab:ablation}), indicating a richness--stability trade-off.
}
\label{tab:register}
\end{table}

Learned embeddings reduce register drift from 3--5/10 to 0/10, and this holds at $K{=}1$ (Table~\ref{tab:register}).
This suggests that register drift is a property of \emph{embedding geometry}---the continuous trajectory crosses register boundaries when the token space is poorly organized---rather than an inherent limitation of the mechanism.
The trade-off is slightly less discourse-rich text, pointing to a richness--stability trade-off mediated by embedding organization; Appendix Table~\ref{tab:fabrication} shows matching reductions in attribution-like patterns and topic drift.

\subsection{Control Surface: Knobs on the Continuous State}
\label{sec:control_surface}

We now ask whether separating prediction from commitment also exposes practically meaningful control knobs.
Table~\ref{tab:knobs} summarizes five that act directly on the evolving continuous state rather than on token probabilities.

\begin{table}[t]
\centering
\small
\setlength{\tabcolsep}{4pt}
\renewcommand{\arraystretch}{1.15}
\begin{tabular}{>{\raggedright\arraybackslash}p{2.3cm}>{\raggedright\arraybackslash}p{2.1cm}>{\raggedright\arraybackslash}p{3.1cm}}
\toprule
\textbf{Control knob} & \textbf{Acts on} & \textbf{Observed effect} \\
\midrule
Direction rate & tail dynamics & discourse richness, convergence speed \\
History noise & prefix state & content diversity without sampling \\
Tail length $K$ & commitment delay & diversity, anti-degeneration \\
State-space CFG & continuous tail & interpretable lookahead \\
Embedding geometry & state manifold & register stability, topic drift \\
\bottomrule
\end{tabular}
\caption{\textbf{Control knobs exposed by continuous-state prediction.} Each knob acts on the evolving state rather than token probabilities.}
\label{tab:knobs}
\end{table}

\textbf{Direction rate controls discourse richness.}
The direction rate $(\eta_{\min}, \eta_{\max})$ governs how aggressively the tail moves toward the model's prediction at each step.
Sweeping from $(0.01, 0.10)$ to $(1.0, 1.0)$ shifts discourse markers from 0.22 to \textbf{0.70} per 100 words at the optimal setting $(0.10, 0.75)$, a $3{\times}$ range (Appendix, Sec.~\ref{sec:rate_sweep}).
Too-slow rates produce incoherent text; too-fast rates eliminate the refinement benefit.

\textbf{History noise provides sampling-free diversity.}
Perturbing committed prefix embeddings with noise $\sigma_h$ yields sampling-free diversity: at $\sigma_h{=}0.5$, content changes substantially across runs while text quality metrics (srep4: .000, d-1: .910) and discourse structure remain stable (Appendix, Sec.~\ref{sec:app_noise}).

\textbf{Embedding geometry controls register stability.}
Learning the embedding space alongside the continuous-prediction adapter reduces register drift from 3--5/10 to 0/10 prompts (Table~\ref{tab:register}), with a corresponding ${\sim}30\%$ reduction in attribution-like patterns (Appendix, Table~\ref{tab:fabrication}), showing that the geometry of the state manifold is itself a control parameter.

%%%%%%%%%%%%%%%%%%%%%%%%%%%%%%%%
% 6. DISCUSSION
%%%%%%%%%%%%%%%%%%%%%%%%%%%%%%%%
\section{Discussion}

\textbf{A different autoregressive interface.}
Standard autoregressive language modeling conflates predicting the next state of generation with committing to a discrete token.
Projected Autoregression separates them by predicting continuously in embedding space and projecting only at commitment time.
The $K{=}1$ decomposition (Table~\ref{tab:ablation}) shows that the regime shift already appears with immediate projection, and extended tail refinement then amplifies it.

\textbf{Scope, generality, and limitations.}
The main study is controlled: same backbone, data, and adaptation budget, with evaluation at 3B LoRA scale on 50 open-ended prompts.
Appendix GPT-2 experiments suggest the phenomenon extends beyond Granite.
Prompt relevance is lower than AR.

\vspace{-4pt}
\section{Conclusion}
Standard autoregressive language modeling fuses prediction with commitment; Projected Autoregression separates them by predicting in embedding space and committing only through projection.
This exposes control variables such as direction rate, history noise, tail length, state-space guidance, and embedding geometry, placing repeated token selection within a broader autoregressive design space.

\section*{Impact Statement}
This paper proposes a reformulation of autoregressive text generation via continuous token dynamics prior to discrete commitment. The primary impact is scientific, potentially informing future decoding and modeling approaches. We do not anticipate societal impacts beyond those commonly associated with language models and their standard deployment risks.

% TODO: no likelihood; scaling; computational cost; dependence on embedding geometry; etc.

%%%%%%%%%%%%%%%%%%%%%%%%%%%%%%%%
% 7. CONCLUSION
%%%%%%%%%%%%%%%%%%%%%%%%%%%%%%%%
% \input{sections/conclusion}

%%%%%%%%%%%%%%%%%%%%%%%%%%%%%%%%
% ACKNOWLEDGEMENTS (optional; remove for blind)
%%%%%%%%%%%%%%%%%%%%%%%%%%%%%%%%
% \section*{Acknowledgements}
% TODO

%%%%%%%%%%%%%%%%%%%%%%%%%%%%%%%%
% REFERENCES
%%%%%%%%%%%%%%%%%%%%%%%%%%%%%%%%
\FloatBarrier
\bibliographystyle{plainnat}

%%%%%%%%%%%%%%%%%%%%%%%%%%%%%%%%
% APPENDIX
%%%%%%%%%%%%%%%%%%%%%%%%%%%%%%%%
\FloatBarrier
\appendix
\clearpage

%%%%%%%%%%%%%%%%%%%%%%%%%%%%%%%%
% APPENDIX — Organized by topic
%%%%%%%%%%%%%%%%%%%%%%%%%%%%%%%%

% ============================================================
% A. TRAINING AND ARCHITECTURE DETAILS
% ============================================================

\section{Architecture}
\label{sec:app_arch}

\begin{figure}[ht!]
\centering
\includegraphics[width=0.85\columnwidth]{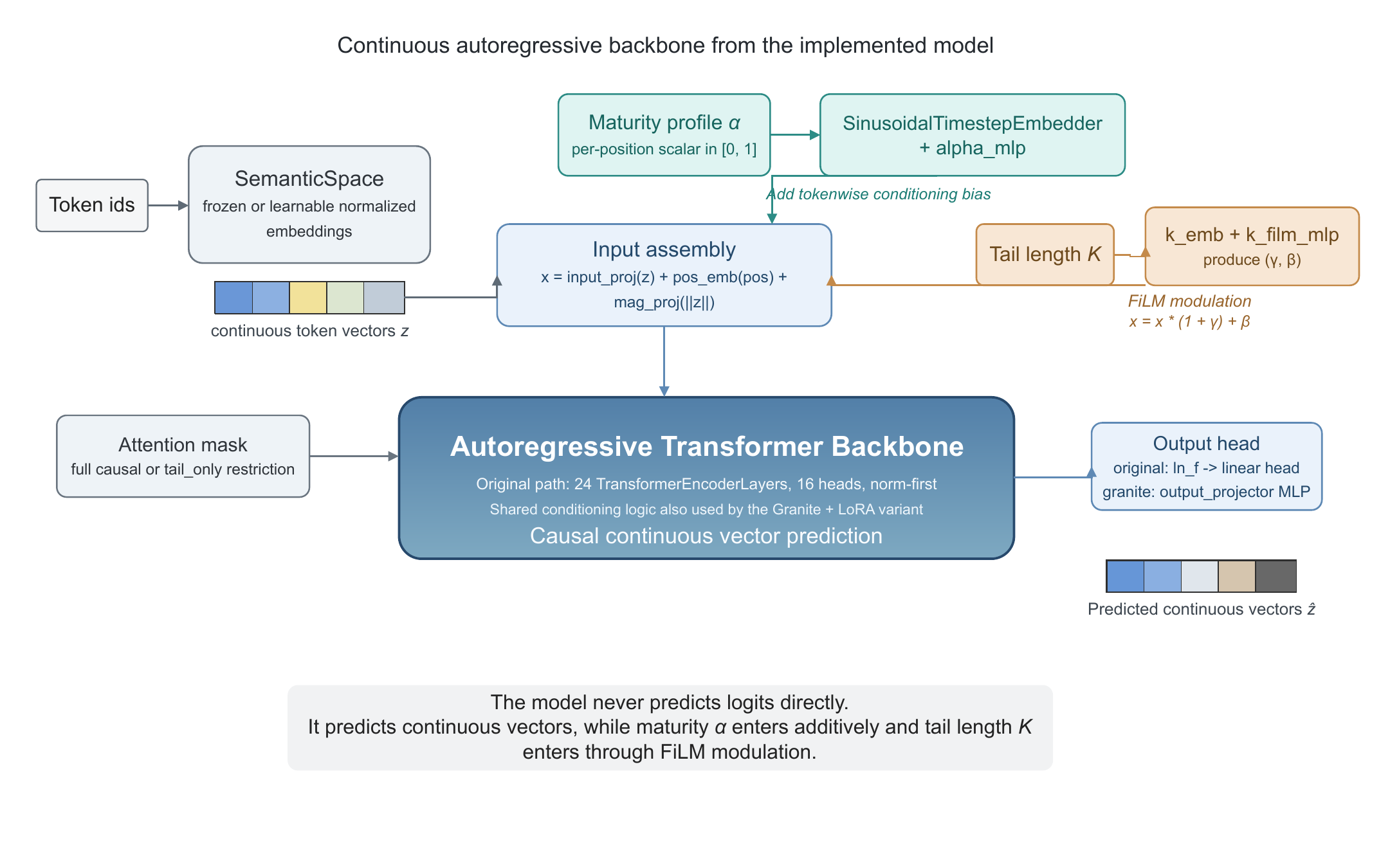}
\caption{
\textbf{Architecture of the Continuous Autoregressive Backbone.}
Input vectors combine a committed prefix with a noisy liquid tail.
The model is conditioned on maturity $\alpha$ (additive bias) and tail length $K$ (FiLM modulation).
}
\label{fig:arch}
\end{figure}

\section{Training Process}
\label{sec:app_training}

\begin{figure*}[t]
\centering
\includegraphics[width=0.9\textwidth]{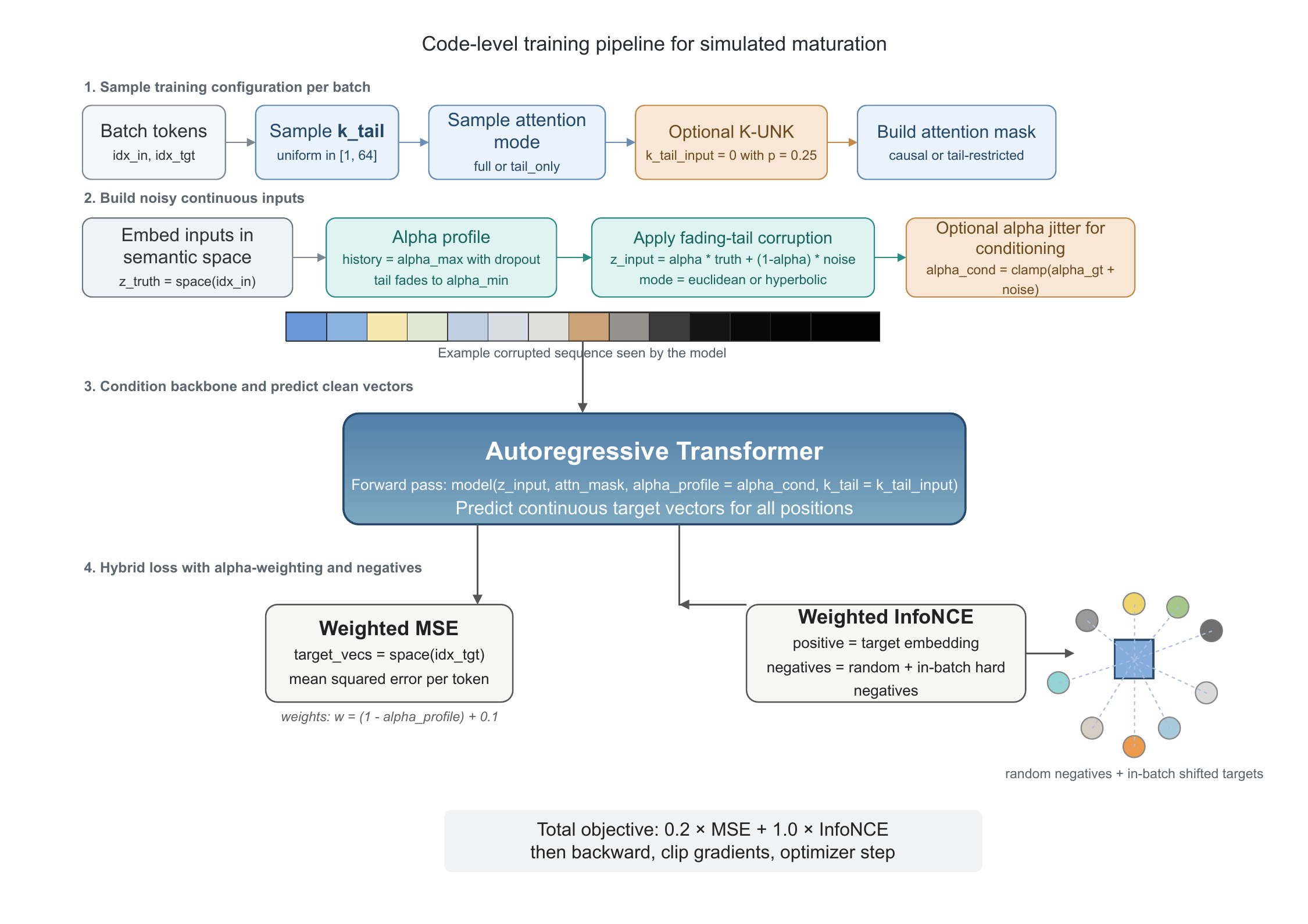}
\caption{
\textbf{Training with Simulated Maturation.}
$\alpha$-dependent noise is injected into tail vectors.
The model is trained with \textbf{MSE Loss} (geometric convergence) and \textbf{InfoNCE Loss} (discriminative structure via negative sampling).
}
\label{fig:training}
\end{figure*}

\paragraph{Alpha profile construction.}
At each training step, a tail length $K$ is sampled uniformly from $\{1, \dots, K_{\max}\}$.
An alpha profile $\boldsymbol{\alpha} \in [0,1]^T$ is constructed with two regions:
(i) the \emph{history} ($t \le T{-}K$), where $\alpha_t = \alpha_{\max}$ (default 0.5) with random dropout (20\% of history tokens are set to $\alpha{=}0$, simulating incomplete visibility);
and (ii) the \emph{tail} ($t > T{-}K$), where $\alpha_t$ fades linearly from $\alpha_{\max}$ to $\alpha_{\min}{=}0$.

\paragraph{Noise injection.}
Ground-truth embeddings are corrupted according to the alpha profile:
\begin{equation}
\tilde{z}_t = \alpha_t \cdot e_{x_t} + (1 - \alpha_t) \cdot \epsilon, \quad \epsilon \sim \mathcal{N}(0, I),
\end{equation}
followed by re-projection to the embedding sphere $\tilde{z}_t \leftarrow R \cdot \tilde{z}_t / \|\tilde{z}_t\|$.
This ensures all vectors---clean or corrupted---lie on the same shell surface, and that magnitude carries no signal about corruption level.

\paragraph{Attention masking.}
Each training step randomly selects one of two attention modes (50/50):
\begin{itemize}[nosep]
  \item \textbf{Full mode:} standard causal mask---every position attends to all preceding positions. This trains the conditioned generation pathway.
  \item \textbf{Tail-only mode:} the causal mask is modified so that tail positions ($t > T{-}K$) are blocked from attending to prefix positions ($t \le T{-}K$). The tail can only attend to itself. This trains the \emph{unconditional} branch, which is the context-masked prediction $\hat{z}^{\text{cm}}$ used in classifier-free guidance at inference (Sec.~\ref{sec:training}).
\end{itemize}
By training with both modes, the model learns to produce meaningful predictions both with and without prefix context, enabling the CFG interpolation $\hat{z} = \hat{z}^{\text{cm}} + s \cdot (\hat{z}^{\text{ca}} - \hat{z}^{\text{cm}})$ at inference.

\paragraph{Tail-length masking ($K_{\text{unk}}$).}
With 25\% probability, the tail length $K$ presented to the model's FiLM conditioning is replaced with a special ``unknown'' token $K_{\text{unk}}$, while the actual noise profile still uses the true $K$.
This prevents the model from relying on exact knowledge of how many positions are liquid and improves robustness when inference uses $K$ values not seen during training.

\paragraph{Alpha jittering.}
When enabled, the alpha values presented to the model's conditioning pathway are jittered: $\alpha_{\text{cond}} = \alpha + \mathcal{N}(0, \sigma_j^2)$, clamped to $[0,1]$.
This prevents the model from relying on the exact maturity value and improves robustness to schedule variations at inference time.

\paragraph{Loss computation.}
The model predicts $\hat{z}_t$ for each position.
The combined loss is $\mathcal{L} = 0.2\,\mathcal{L}_{\text{MSE}} + 1.0\,\mathcal{L}_{\text{NCE}}$, both weighted by $(1 - \alpha_t + 0.1)$ so that positions near commitment (low $\alpha$) receive higher gradient signal.
The NCE loss uses a mixture of random negatives (sampled uniformly from the vocabulary) and \emph{hard in-batch negatives} (shifted target tokens from the same batch), with a configurable ratio (default 50\% hard).
Logits are scaled by a temperature parameter $\tau$ before the cross-entropy.

\paragraph{Optimization.}
We use AdamW with separate learning rates for LoRA parameters and (optionally) embedding parameters.
Gradient accumulation is used to reach effective batch sizes on single-GPU setups (batch size 1 with 16 accumulation steps).
Gradients are clipped to norm 1.0 for both the model and the embedding space.
Training runs for 50K--600K steps depending on the embedding configuration.

% ============================================================
% B. PARAMETER SENSITIVITY
% ============================================================

\section{Parameter Sensitivity}
\label{sec:app_sensitivity}

\subsection{Direction Rate}
\label{sec:rate_sweep}

The direction rate $(\eta_{\min}, \eta_{\max})$ controls how aggressively the tail moves toward the model's prediction at each refinement step.
At each step, tail position $t$ is updated as $z_t \leftarrow z_t + \eta_t \cdot (\hat{z}_t - z_t)$, where $\eta_t$ is linearly interpolated between $\eta_{\min}$ (for the most mature tail position) and $\eta_{\max}$ (for the least mature).
Low rates produce slow, cautious convergence; high rates produce aggressive, nearly immediate jumps to the model's prediction.

We sweep six configurations from $(0.01, 0.10)$ to $(1.0, 1.0)$.
The rate $(1.0, 1.0)$ is a special case where the tail vector is replaced entirely by the model's prediction at every step, eliminating the gradual convergence that characterizes the maturation process.

\begin{table}[ht!]
\centering
\small
\setlength{\tabcolsep}{3pt}
\renewcommand{\arraystretch}{1.15}
\begin{tabular}{lccccc}
\toprule
\textbf{Rate} $(\eta_{\min}, \eta_{\max})$ & \textbf{len} & \textbf{srep4} & \textbf{d-1} & \textbf{disc.} & \textbf{info-d} \\
\midrule
(0.01, 0.10) & 46 & .000 & .986 & 0.22 & 13.2$^*$ \\
(0.05, 0.30) & 67 & .000 & .935 & 0.41 & 7.9 \\
\textbf{(0.10, 0.75)} & \textbf{86} & \textbf{.000} & \textbf{.916} & \textbf{0.70} & \textbf{11.7} \\
(0.30, 0.75) & 96 & .002 & .888 & 0.53 & 12.9 \\
(0.50, 0.90) & 82 & .000 & .922 & 0.20 & 11.0 \\
(1.00, 1.00) & 84 & .000 & .912 & 0.29 & 11.5 \\
\bottomrule
\end{tabular}
\caption{Direction rate sweep. The default $(0.10, 0.75)$ maximizes discourse markers. Too-slow rates ($< 0.05$) produce incoherent text ($^*$artifacts). $K{=}16$, 10 prompts.}
\label{tab:rates}
\end{table}

Table~\ref{tab:rates} reveals a clear sweet spot.
Too-slow rates ($\eta_{\max} \le 0.10$) prevent the tail from converging within the available refinement steps, producing short outputs (46 words) with artificially high diversity (d-1~$= .986$) due to incoherent token sequences.
The high information density score (13.2) for the slowest rate is an artifact: the short, incoherent output happens to contain many unique content words.
Too-fast rates ($\eta_{\max} \ge 0.90$) eliminate the gradual refinement process entirely, reducing discourse richness (disc.~$= 0.20$) as the model loses the benefit of iterative neighborhood narrowing.

The default rate $(0.10, 0.75)$ balances convergence speed with refinement depth: the tail converges fast enough to produce coherent text but slowly enough to explore the semantic landscape, resulting in the highest discourse marker density (0.70/100w) and strong information density (11.7).

\subsection{Tail Length $K$}
\label{sec:k_sweep}

We evaluate $K \in \{2, 4, 8, 16, 32\}$ with and without warmup (pre-converging the tail before committing).
Surprisingly, text quality metrics remain similar across all $K$ values: all configurations produce near-zero repetition (srep4~$\le$~.002), high lexical diversity (d-1~$\ge$~.88), and comparable information density.
This is consistent with the mechanistic finding that neighborhood narrowing saturates at $K \approx 16$ (Sec.~\ref{sec:mechanistic}): beyond this point, additional refinement steps yield diminishing returns because the tail has already converged to a small neighborhood of plausible tokens.

A subtle factor moderates the effect of $K$: the \emph{growing tail}.
At the start of generation, the tail has fewer than $K$ positions available (e.g., after committing 3 tokens, only 3 tail positions exist even if $K{=}32$).
This means the effective tail length starts at 1 and grows until it reaches $K$, diluting the average impact of large $K$ values over the full generation.

\paragraph{Warmup.}
We also evaluate a \emph{warmup} variant in which the tail is pre-converged (multiple refinement cycles) before the first commitment.
This ensures the tail reaches its intended length and has undergone full neighborhood narrowing before any tokens are committed.
Warmup has minimal effect at small $K$ but matters substantially at $K{=}32$, where output length increases from 75 to 108 words.
Without warmup, the first few committed tokens are produced from a partially-formed tail, leading to early EOS commitment; with warmup, the tail has time to stabilize and the model produces longer, more coherent outputs.

\subsection{History Noise}
\label{sec:app_noise}

\paragraph{Rationale.}
In standard Projected Autoregression, committed prefix vectors are fed back to the model exactly as they were at commitment time.
This makes generation fully deterministic given the initial random embryo: the same embryo always produces the same output.
History noise ($\sigma_h$) introduces controlled stochasticity by perturbing committed prefix embeddings before they are passed to the model at each refinement step.
The key insight is that this perturbation acts as \emph{simulated annealing} over the generation trajectory: the noise perturbs the optimization landscape that the tail navigates, causing the tail to explore different regions of embedding space, without affecting the deterministic commitment rule itself (which remains argmax over cosine similarity).

Concretely, history noise sets $\alpha_{\text{history}} = 1 - \sigma_h$ for committed positions, so the prefix vectors seen by the model become:
\begin{equation}
\tilde{z}_t^{\text{prefix}} = (1 - \sigma_h) \cdot z_t + \sigma_h \cdot \epsilon, \quad \epsilon \sim \mathcal{N}(0, I),
\end{equation}
followed by re-projection to the embedding sphere.
At $\sigma_h{=}0$, the prefix is exact and generation is deterministic.
At $\sigma_h{=}0.5$, each prefix vector is a 50/50 mix of the true embedding and random noise---a substantial perturbation that nonetheless preserves the general direction of the vector.

\paragraph{Effect on generation.}
History noise diversifies \emph{content} while preserving \emph{structure}.
Table~\ref{tab:hist_noise} shows that text quality metrics are robust up to $\sigma_h{=}0.5$: repetition remains near zero, lexical diversity stays high, and the overall discourse structure is maintained.
Beyond $\sigma_h{=}0.5$, the prefix becomes too corrupted for the model to maintain coherence.

\begin{table}[ht!]
\centering
\small
\begin{tabular}{rccc}
\toprule
$\sigma_h$ & \textbf{srep4}$\downarrow$ & \textbf{d-1}$\uparrow$ & \textbf{Effect} \\
\midrule
0.0 & .001 & .911 & Deterministic baseline \\
0.2 & .000 & .912 & Mild content variation \\
\textbf{0.5} & \textbf{.000} & \textbf{.910} & Different content, same structure \\
\bottomrule
\end{tabular}
\caption{History noise preserves text quality metrics up to $\sigma_h{=}0.5$.}
\label{tab:hist_noise}
\end{table}

\paragraph{Qualitative examples.}
The effect is best illustrated through paired examples.
With the prompt \emph{``In the summer of 1969, humanity achieved something remarkable when''}:

\begin{itemize}[nosep]
  \item \textbf{Without noise} ($\sigma_h{=}0$): the model generates text about biology---``it discovered that there were two different types of life on Earth---a small group called `bacteria' and another large group known as `eukaryotes.'\,''
  \item \textbf{With noise} ($\sigma_h{=}0.5$): the model generates text about space technology---``it launched its first satellite into space---the SATELITE~A~(SA). This pioneering project showed us that we could send messages across the globe\ldots''
\end{itemize}

Both outputs maintain the same expository register, similar sentence structure, and comparable information density---but the \emph{topic} shifts entirely.
Similarly, for the prompt \emph{``When the first settlers arrived in the valley, they found''}:

\begin{itemize}[nosep]
  \item \textbf{Without noise:} generates text about Indian civilizations---``these small groups evolved into larger societies called `civilizations,' which developed their own language (Maharashtra Linga), religion (Hinduism), and traditions.''
  \item \textbf{With noise:} generates text about Mexico---``This is why we call this part of Mexico `the Grand Canyon'\,''---a completely different geographic and cultural context.
\end{itemize}

\paragraph{Interpretation.}
This behavior is consistent with the centroid-convergence mechanism described in Sec.~\ref{sec:mechanistic}.
The tail converges to the centroid of a neighborhood of plausible tokens; perturbing the prefix shifts \emph{which} neighborhood the tail converges to, but the convergence dynamics themselves remain stable.
In the annealing analogy: the noise changes the energy landscape (by presenting a slightly different context), causing the system to settle into a different local minimum, but the overall optimization procedure---and thus the structural properties of the output---remains unchanged.
This provides a principled diversity control knob that operates in continuous space, orthogonal to token-level sampling heuristics like temperature or top-$p$.

\subsection{Determinism vs.\ Tail Length}
\label{sec:app_determinism}

Projected Autoregression uses deterministic argmax decoding: given a fixed state, the committed token is always the vocabulary item with highest cosine similarity to the predicted vector.
Yet the system produces different outputs across runs because the \emph{embryo}---the initial random vectors that fill the liquid tail---is sampled fresh each time.
This creates a unique form of controlled stochasticity: diversity arises from initialization rather than from sampling at the decision boundary.

We measure run-to-run consistency via pairwise Jaccard similarity over unigrams (5 independent runs per $K$, averaged over 10 prompts).
Higher Jaccard indicates more deterministic behavior (identical runs produce similar token sets).

\begin{table}[ht!]
\centering
\small
\begin{tabular}{rcc}
\toprule
$K$ & Avg.\ length (words) & Jaccard $\uparrow$ \\
\midrule
2 & 115 & 0.213 \\
4 & 119 & 0.196 \\
8 & 114 & 0.197 \\
16 & 114 & 0.207 \\
32 & 114 & 0.219 \\
\textbf{64} & \textbf{102} & \textbf{0.443} \\
128 & 106 & 0.331 \\
\bottomrule
\end{tabular}
\caption{Determinism peaks at $K{\approx}64$: two competing forces---convergence (increases with $K$) vs.\ compound noise from random tail neighbors (also increases with $K$).}
\label{tab:determinism}
\end{table}

Table~\ref{tab:determinism} reveals a non-monotonic relationship between $K$ and determinism, peaking at $K{\approx}64$ (Jaccard~$= 0.443$).
This reflects two competing forces:
\begin{enumerate}[nosep]
  \item \textbf{Convergence force:} larger $K$ means more refinement steps before commitment, allowing the tail to converge more thoroughly toward a stable neighborhood. This increases determinism.
  \item \textbf{Compound noise force:} larger $K$ means more random tail vectors interacting through self-attention at each step. Each random neighbor contributes a small perturbation to the predictions for other tail positions. Over many positions, these perturbations compound and can push the trajectory into different basins of attraction. This decreases determinism.
\end{enumerate}
At $K{\approx}64$, convergence dominates: the tail has enough steps to overcome initialization noise, and the random perturbations have not yet compounded sufficiently to destabilize the trajectory.
Beyond $K{=}64$, the compound noise from 128 random tail vectors overwhelms the convergence benefit, and determinism drops back down.
Note that output length remains stable across $K$ values (102--119 words), confirming that the determinism variation reflects genuine content diversity rather than length artifacts.

% ============================================================
% C. TEXT QUALITY ANALYSIS (EXTENDED)
% ============================================================

\section{Extended Text Analysis}

\subsection{Fuzzy Associative Retrieval}
\label{sec:fuzzy_retrieval}

A distinctive feature of Projected Autoregression is that its ``hallucinations'' are qualitatively different from those of standard AR models.
Rather than producing arbitrary confabulations, the model generates claims that are \emph{approximately correct}---real entities with wrong dates, real institutions with wrong founding years, real people in slightly wrong roles.
We term this \emph{fuzzy associative retrieval}: the model retrieves genuine knowledge from its parameters but recombines it imprecisely.

This pattern is a natural consequence of the centroid-convergence mechanism (Sec.~\ref{sec:mechanistic}).
The liquid tail converges to the centroid of a neighborhood of plausible tokens, not to any single token embedding.
When the neighborhood contains multiple semantically related entities (e.g., several lighthouse engineers, several founding dates), the centroid represents a ``blend'' of these entities.
Upon projection to discrete tokens, this blend resolves to whichever entity's embedding is closest---which may not be the factually correct one for the given context.

To quantify this phenomenon, we manually verified five representative claims from generated outputs against external sources.

\begin{table}[ht!]
\centering
\small
\setlength{\tabcolsep}{3pt}
\renewcommand{\arraystretch}{1.15}
\begin{tabular}{p{3.8cm}p{3.2cm}c}
\toprule
\textbf{Generated claim} & \textbf{Verified fact} & \textbf{Status} \\
\midrule
1969: bacteria vs.\ eukaryotes & Whittaker (1969): five-kingdom classification & \cmark \\
John Bowen, lighthouse, 1846 & Sir John Bowen, lighthouse engineer, 1886 & \cmark \\
Grand Rapids Public Library, 1867 & Real library, founded 1871 & \cmark \\
David Hayes, director & Emmy-winning director (Netflix) & \cmark \\
Bleach Cultural Conservancy & No match found & \xmark \\
\bottomrule
\end{tabular}
\caption{Fuzzy retrieval verification.
Outputs are predominantly grounded in real knowledge but recombined imprecisely.}
\label{tab:fuzzy}
\end{table}

Four of five claims are grounded in real entities, with errors confined to peripheral details (dates, specific roles).
For example, the model generates ``John Bowen, lighthouse, 1846'' when the real Sir John Bowen was a lighthouse engineer active in 1886---the correct person and profession, but with a 40-year date shift.
Similarly, ``Grand Rapids Public Library, 1867'' refers to a real institution founded in 1871.
These errors are consistent with centroid-based misbinding: the model retrieves the correct semantic neighborhood (lighthouse engineers, library founding dates) but commits to a slightly wrong point within that neighborhood.

The one complete fabrication (``Bleach Cultural Conservancy'') is notable precisely because it is rare.
This suggests that the continuous prediction mechanism primarily \emph{retrieves} from the model's stored knowledge rather than generating arbitrary text, but does so through a geometric operation (centroid convergence) that introduces systematic imprecision.
This is qualitatively different from standard AR hallucination, where factual errors typically arise from token-by-token sequential probability without the geometric blending that characterizes Projected Autoregression.

\subsection{Attribution-Like Patterns and Topic Drift}
\label{sec:app_fabrication}

Two systematic text quality issues emerge across generation methods: \emph{attribution-like patterns} and \emph{topic drift}.

Attribution-like patterns are constructions of the form ``Dr.~X, director of Y'' or ``according to Z, professor of W''---fabricated expert citations that lend false authority to generated text.
These arise naturally in continuous-space generation because the centroid of a semantic neighborhood often lies near ``expert + institution'' templates, which are common structural patterns in the FineWeb training data.

Topic drift measures how much the content of the generated text shifts between its beginning and end.
We compute it as $1 - J$, where $J$ is the Jaccard overlap between content words (nouns, verbs, adjectives) in the first and last thirds of the output.
High topic drift indicates that the generation has wandered far from its initial subject---a common failure mode in long-form generation.

We automatically measure both metrics across all methods (50 prompts) using lexical template matching for attributions and Jaccard overlap for drift.

\begin{table}[ht!]
\centering
\small
\setlength{\tabcolsep}{3pt}
\renewcommand{\arraystretch}{1.15}
\begin{tabular}{lcc}
\toprule
\textbf{Method} & \textbf{Attrib./100w}$\downarrow$ & \textbf{Topic drift}$\downarrow$ \\
\midrule
Frozen $K{=}1$ & 0.53 & 0.930 \\
Frozen $K{=}16$ & 0.48 & 0.952 \\
Learned $K{=}1$ & 0.36 & \textbf{0.884} \\
Learned $K{=}16$ & \textbf{0.33} & 0.916 \\
\midrule
AR greedy & 0.09 & 0.454 \\
AR greedy+rep & 0.44 & 0.966 \\
AR top-$p$=0.95 & 0.14 & 0.906 \\
\bottomrule
\end{tabular}
\caption{Attribution-like patterns and topic drift.
AR with repetition penalty produces attribution patterns at a similar rate to frozen Proj.\ AR (0.44 vs.\ 0.48).
AR greedy shows 0\%, likely because its dominant failure mode is repetition rather than template drift.
Learned embeddings reduce both.}
\label{tab:fabrication}
\end{table}

Several patterns emerge from Table~\ref{tab:fabrication}.
First, AR greedy shows near-zero attribution patterns (0.09/100w) because its dominant failure mode is repetition, not template drift---the model gets stuck in loops before it can produce attribution constructions.
AR with repetition penalty (0.44/100w) produces attribution patterns at a rate comparable to frozen Projected AR (0.48/100w), suggesting that once repetition is suppressed, both methods are similarly prone to template attractors in the frozen embedding space.

Second, learned embeddings reduce attribution patterns by ${\sim}30\%$ (from 0.48 to 0.33 at $K{=}16$) and also reduce topic drift.
This is consistent with the embedding geometry analysis (Sec.~\ref{sec:exp_embeddings}): learned embeddings reorganize the token space so that semantic neighborhoods are better structured, reducing the likelihood that centroid convergence lands on attribution-like templates.

Third, topic drift is high across all methods ($>0.88$), reflecting the inherent difficulty of maintaining topical coherence over 100+ tokens of generation.
AR greedy is an outlier (0.454) only because its repetitive outputs trivially preserve the same content words throughout.

% Qualitative comparison is included in Cross-Backbone Evidence (Sec.~\ref{sec:app_gpt2})

% ============================================================
% E. EMBEDDING GEOMETRY
% ============================================================

\section{Embedding Geometry}
\label{sec:exp_embeddings}

The geometry of the embedding space plays a critical role in Projected Autoregression because the commitment operation (nearest-neighbor projection) depends directly on how tokens are arranged on the embedding sphere.
When embeddings are fine-tuned during training, the model can reorganize this space to better support continuous-state generation.

\begin{figure}[ht!]
  \centering
  \includegraphics[width=\linewidth]{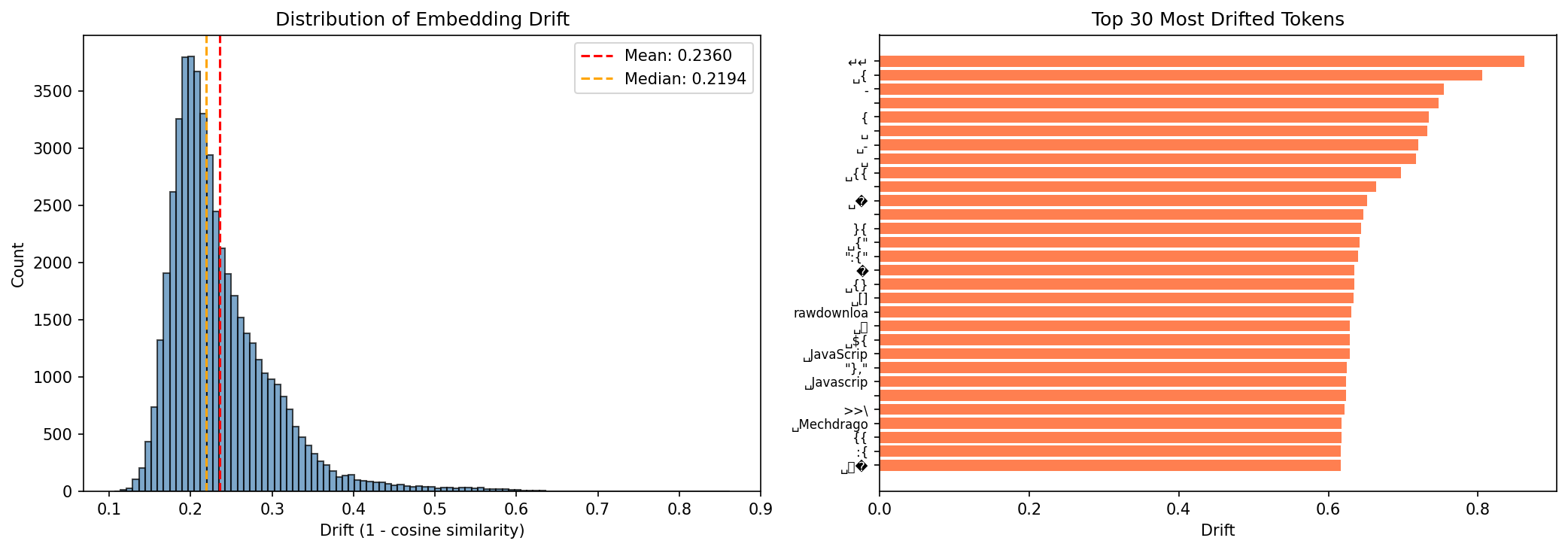}
  \caption{Embedding drift from frozen GPT-2 to learned embeddings.}
  \label{fig:embedding_drift}
\end{figure}

Figure~\ref{fig:embedding_drift} visualizes the drift between frozen GPT-2 embeddings and learned embeddings after 600K training steps.
We compute drift as the cosine distance between each token's original and learned embedding vector.

\begin{itemize}[nosep]
  \item \textbf{Stable tokens:} Years (``1990'', ``2005'') and high-frequency function words (``the'', ``of'', ``and'') exhibit minimal drift. These tokens already occupy well-defined positions in the pretrained space that serve the continuous prediction objective well.
  \item \textbf{High-drift tokens:} Punctuation marks and rare symbols drift substantially. These tokens have arbitrary positions in the pretrained space (determined by subword statistics rather than semantic content) and are repositioned by the training objective to better support centroid-based prediction.
  \item \textbf{Semantic reorganization:} Content words undergo meaningful reorganization. For example, ``Python'' moves from proximity to other programming \emph{languages} (Java, PHP, Ruby) toward programming \emph{culture} tokens (Lisp, Emacs, Unix). This suggests the model learns an embedding organization optimized for the semantic neighborhoods that arise during centroid convergence, rather than the distributional similarity that organizes pretrained embeddings.
\end{itemize}

\paragraph{Impact on neighborhood structure.}
The practical consequence of this reorganization is that nearest-neighbor neighborhoods become more semantically coherent.
With frozen embeddings, the neighbors of a given token are often subword artifacts (tokens that share character sequences but not meaning).
With learned embeddings, neighbors tend to be semantically related words.
For instance, the neighborhood of the token for ``write'' in learned embeddings contains ``print'', ``produce'', ``publish''---words that are semantically substitutable in many contexts.
This improved neighborhood structure directly benefits the centroid-convergence mechanism: when the tail converges to a neighborhood centroid, that centroid is more likely to represent a coherent semantic concept rather than a blend of subword noise.
This explains why learned embeddings reduce register drift (Table~\ref{tab:register}) and attribution-like patterns (Table~\ref{tab:fabrication}).

% ============================================================
% F. CLASSIFIER-FREE GUIDANCE
% ============================================================

\section{Classifier-Free Guidance}
\label{sec:exp_cfg}

Classifier-free guidance (CFG) is a technique originally developed for diffusion models~\cite{ho2022classifier} that we adapt to Projected Autoregression.
At each refinement step, the model produces two predictions: a \emph{conditional} prediction $\hat{z}^{\text{ca}}$ (using the full causal mask, attending to the prefix) and an \emph{unconditional} prediction $\hat{z}^{\text{cm}}$ (using the tail-only mask, where tail positions cannot attend to the prefix).
The final prediction is an extrapolation:
\begin{equation}
\hat{z} = \hat{z}^{\text{cm}} + s \cdot (\hat{z}^{\text{ca}} - \hat{z}^{\text{cm}}),
\end{equation}
where $s$ is the guidance scale.
At $s{=}1$, this reduces to the conditional prediction.
At $s{>}1$, the model ``overshoots'' in the direction of context-dependence, amplifying features that distinguish the conditional prediction from the unconditional one.

The unconditional branch is trained via the tail-only attention mode described in Sec.~\ref{sec:app_training}: with 50\% probability during training, the attention mask blocks tail positions from attending to the prefix, forcing the model to predict tail vectors purely from tail-internal context.
This ensures both branches produce well-calibrated predictions.

\begin{figure*}[ht!]
  \centering
  \includegraphics[width=\linewidth]{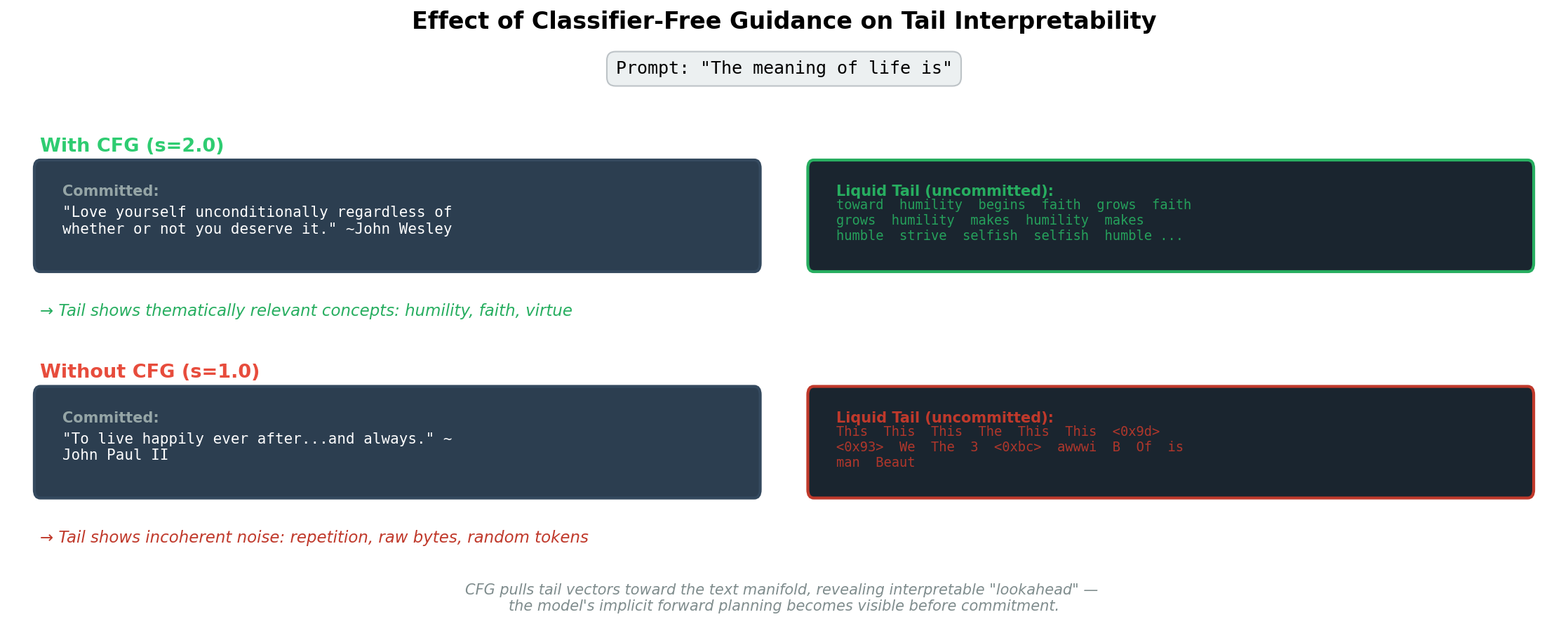}
  \caption{Effect of CFG on tail interpretability.
  Without CFG (top): incoherent noise.
  With CFG (bottom): semantically meaningful lookahead.}
  \label{fig:cfg_comparison}
\end{figure*}

Figure~\ref{fig:cfg_comparison} shows the qualitative effect of CFG on tail interpretability.
Without guidance ($s{=}0$), the tail vectors project to incoherent vocabulary items that bear no semantic relationship to the prefix or to each other.
With guidance ($s{=}2$), the same tail positions project to tokens that form coherent lookahead sequences---semantically meaningful continuations of the committed prefix.
In our experiments, guidance scales in the range $s \in [1.5, 3.0]$ produce the best results, with higher values increasing the coherence of the tail at the cost of reduced diversity.

Whether the coherent tail states under CFG reflect genuine forward planning (the model ``looking ahead'' at future tokens) or simply coherent interpolation in embedding space (the extrapolation happening to land in semantically consistent regions) remains an open question.
Both explanations are consistent with the observed behavior, and distinguishing them would require probing the model's internal representations more deeply.

% ============================================================
% G. CROSS-BACKBONE EVIDENCE
% ============================================================

\section{Cross-Backbone Evidence (GPT-2)}
\label{sec:app_gpt2}

To verify that the phenomena described in this paper are not specific to the Granite~3B backbone, we replicate the core experiments on GPT-2 Medium (${\sim}$350M parameters) with learned embeddings trained for 600K steps.
This smaller model allows us to examine whether the continuous prediction regime, neighborhood narrowing, and resistance to repetition degeneration emerge from the \emph{framework} itself rather than from the specific backbone architecture or scale.

\begin{figure*}[t]
  \centering
  \includegraphics[width=0.95\textwidth]{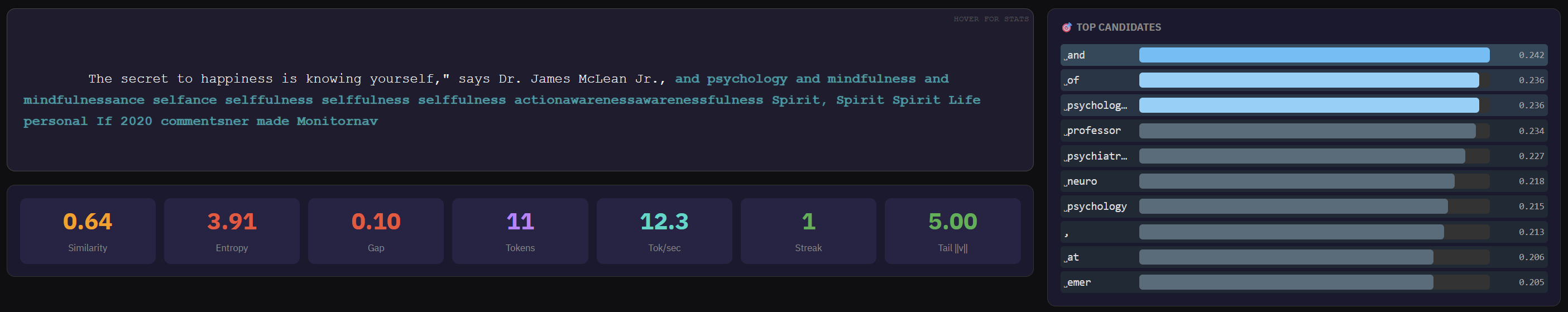}
  \caption{
  Generation interface showing Projected Autoregression on GPT-2 Medium.
  Committed text (white) followed by the liquid tail (cyan).
  Despite high entropy ($H{=}3.91$), all top candidates are contextually relevant.
  }
  \label{fig:qualitative_ui}
\end{figure*}

Figure~\ref{fig:qualitative_ui} shows the generation interface on GPT-2 Medium.
The committed prefix (white) is followed by the liquid tail (cyan), with the top-$k$ nearest vocabulary tokens displayed for each tail position.
Despite high predictive entropy ($H{=}3.91$ nats over the full vocabulary), all top candidates at each tail position are contextually relevant---consistent with the neighborhood narrowing mechanism observed on Granite~3B.

The qualitative comparisons and trajectory analyses that follow demonstrate that Projected Autoregression maintains coherent, non-degenerate generation under deterministic argmax decoding even on this smaller backbone, while standard GPT-2 greedy decoding collapses into repetition under identical conditions.

\subsection{Qualitative Comparison: Stability Under Deterministic Decoding}
\label{app:samples}

We qualitatively compare the generative stability of \textbf{Projected Autoregression} with a standard autoregressive baseline (\textbf{GPT-2 Medium}).
Both models share the same architectural backbone (24-layer Transformer, $\sim$350M parameters), enabling a controlled comparison under comparable computational constraints.

Standard autoregressive language models rely on immediate discrete commitment at each generation step, which induces a well-known trade-off: greedy decoding often collapses into repetition, while sampling-based methods trade determinism for coherence.
In contrast, Projected Autoregression resolves uncertainty through continuous token dynamics prior to discretization.
The table below illustrates that under fully deterministic conditions, standard greedy decoding collapses, whereas Projected Autoregression maintains coherent generation.

\begin{longtable}{p{0.12\textwidth} p{0.85\textwidth}}
\toprule
\textbf{Task} & \textbf{Side-by-Side Comparison} \\
\midrule
\endhead

% --- DUGMA 1: Repetition Stress Test ---
\textbf{Repetition Stress Test} &
\textbf{Prompt:} I went to the store to buy some apples, oranges, bananas, and \\
& \vspace{-5pt}
\begin{minipage}[t]{0.48\linewidth}
    \small
    \textbf{\textcolor{red}{\ding{55} Standard AR Baseline (GPT-2, Greedy)}} \\
    ...grapes. I was so excited to see the store, but I was so nervous. I was so nervous that I didn't know if I would be able to buy anything. I was so nervous that I didn't know if I would be able to buy anything. \textcolor{red}{\textbf{[Repetition Loop]}}
\end{minipage}
\hfill
\vline
\hfill
\begin{minipage}[t]{0.48\linewidth}
    \small
    \textbf{\textcolor{teal}{\ding{51} Projected Autoregression (Ours, Argmax)}} \\
    ...cherries. They were packed nicely inside plastic bags with labels attached saying `Made By Organic Plant Based Ingredients'. These weren't available anywhere else until after Christmas Eve when I saw them at Walmart for \$5 each!!
\end{minipage} \\
\midrule

% --- DUGMA 2: Narrative ---
\textbf{Narrative Continuation} &
\textbf{Prompt:} The fog rolled into the harbor like a thick blanket, covering everything in silence. \\
& \vspace{-5pt}
\begin{minipage}[t]{0.48\linewidth}
    \small
    \textbf{\textcolor{red}{\ding{55} Standard AR Baseline (GPT-2, Greedy)}} \\
    ..."I'm sorry, I'm sorry, I'm sorry, I'm sorry, I'm sorry, I'm sorry, I'm sorry, I'm sorry, I'm sorry, I'm sorry, I'm sorry, I'm sorry, I'm sorry, I... \textcolor{red}{\textbf{[Collapse]}}
\end{minipage}
\hfill
\vline
\hfill
\begin{minipage}[t]{0.48\linewidth}
    \small
    \textbf{\textcolor{teal}{\ding{51} Projected Autoregression (Ours, Argmax)}} \\
    ...There was no sound except thunder coming from beneath the waves crashing against the rocks along the shoreline. As they reached us, we saw white streaks forming around them; then suddenly, dark bodies appeared floating at our feet.
\end{minipage} \\
\midrule

% --- DUGMA 3: Dialogue ---
\textbf{Dialogue Consistency} &
\textbf{Prompt:} "I can't believe you sold the car without asking me!" Sarah shouted. \\
& \vspace{-5pt}
\begin{minipage}[t]{0.48\linewidth}
    \small
    \textbf{\textcolor{red}{\ding{55} Standard AR Baseline (GPT-2, Greedy)}} \\
    ..."I'm sorry, I didn't know you were going to be here," I said. "I'm sorry, I didn't know you were going to be here." "I'm sorry, I didn't know you were going to be here," Sarah said again. \textcolor{red}{\textbf{[Repetition Loop]}}
\end{minipage}
\hfill
\vline
\hfill
\begin{minipage}[t]{0.48\linewidth}
    \small
    \textbf{\textcolor{teal}{\ding{51} Projected Autoregression (Ours, Argmax)}} \\
    ..."You're not kidding yourself," she replied angrily. Her voice sounded distant, as if she were replaying the moment in her head. Perhaps she already knew what was coming next, or perhaps she was simply too tired to argue further.
\end{minipage} \\
\midrule

% --- DUGMA 4: Scientific ---
\textbf{Expository Explanation} &
\textbf{Prompt:} The central dogma of molecular biology describes \\
& \vspace{-5pt}
\begin{minipage}[t]{0.48\linewidth}
    \small
    \textbf{\textcolor{red}{\ding{55} Standard AR Baseline (GPT-2, Greedy)}} \\
    ...the process of evolution by natural selection. The theory of evolution is based on the idea that organisms are designed by God to survive and reproduce. The theory of evolution is based on the idea that organisms are designed by God... \textcolor{red}{\textbf{[Repetition \& Factual Error]}}
\end{minipage}
\hfill
\vline
\hfill
\begin{minipage}[t]{0.48\linewidth}
    \small
    \textbf{\textcolor{teal}{\ding{51} Projected Autoregression (Ours, Argmax)}} \\
    ...the flow of genetic information from DNA to RNA and proteins within biological systems. This framework describes how genetic instructions are transcribed and translated to regulate cellular structure, function, and replication.
\end{minipage} \\
\bottomrule
\caption{\textbf{Qualitative Stability Under Deterministic Decoding.}
Both models use fully deterministic greedy decoding (argmax).
The standard autoregressive baseline collapses into repetition or produces incoherent or factually incorrect continuations.
In contrast, Projected Autoregression maintains stable and coherent generation through continuous token dynamics, without requiring stochastic sampling or heuristic penalties.}
\label{tab:comparison}
\end{longtable}

\subsection{Maturation Trajectory Statistics}
\label{app:dynamics}

To understand the internal behavior of the liquid tail on a smaller backbone, we analyzed the maturation trajectories of randomly selected tokens during GPT-2 generation ($n=15$, tail length $K=16$).
Table~\ref{tab:dynamics} summarizes the key metrics characterizing how tokens evolve from initialization to final commitment.
These statistics are consistent with the neighborhood narrowing analysis on Granite~3B (Sec.~\ref{sec:mechanistic}): both backbones show late convergence and active exploration before commitment.

\begin{table}[ht!]
\centering
\small
\renewcommand{\arraystretch}{1.2}
\begin{tabular}{l c p{8cm}}
\toprule
\textbf{Metric} & \textbf{Value} & \textbf{Interpretation} \\
\midrule
\textbf{Birth Accuracy} & 0\% & Tokens initialize in an incorrect state. The maturation process is strictly necessary for correct prediction---the model cannot ``skip'' the tail. \\
\midrule
\textbf{Path Efficiency} & 51\% & Trajectories are not linear interpolations. Low efficiency indicates curved paths in embedding space, suggesting the model explores intermediate semantic concepts before converging. \\
\midrule
\textbf{Avg.\ Convergence} & 14.4 / 16 & Tokens stabilize late in the maturation window (step $\sim$14 of 16), indicating full utilization of the lookahead capacity. \\
\midrule
\textbf{Unique Top-1} & 8.0 & On average, a token changes its top candidate identity 8 times during maturation, explaining the model's resistance to repetition loops. \\
\midrule
\textbf{Final Confidence} & 0.21 & The model commits to a token despite low predictive probability (21\%), relying on geometric stability rather than softmax certainty. \\
\bottomrule
\end{tabular}
\caption{\textbf{Trajectory Statistics (GPT-2 Medium).} Metrics computed over maturing tokens show active exploration, late convergence, and commitment based on geometric position rather than probability concentration.}
\label{tab:dynamics}
\end{table}

These statistics reinforce the framework's key properties across backbones:
(1)~the 0\% birth accuracy and 51\% path efficiency confirm non-trivial computation through curved trajectories;
(2)~convergence at step 14.4 of 16 shows the model maintains plasticity as long as possible;
(3)~low final confidence (0.21) demonstrates that commitment occurs without probability collapse, consistent with the neighborhood narrowing mechanism on Granite~3B where the tail converges to a centroid of plausible tokens rather than to a single high-confidence prediction.

\subsection{Syntactic Stability Under Extended Tail}
\label{app:syntax}

We analyze a GPT-2 generation sample produced with an extended liquid tail ($K=32$) to examine the effect of token maturation on long-range structural coherence.
This setting probes a known failure mode of greedy decoding in small autoregressive language models, where syntactic collapse or repetition often occurs in long, nested constructions.

\textbf{Generated Sample:}
\begin{quote}
\textit{``...Authorities say they believe there may be more than one person involved in this incident but no further information has been released yet regarding any injuries or fatalities among those who were on board that plane which was heading south from New York City at the time and also took off for a trip back into Canada...''}
\end{quote}

The model maintains grammatical structure across a sentence exceeding 50 tokens, successfully tracking multiple nested clauses (relative clauses, coordinated verb phrases) without repetition or syntactic collapse.
Content inconsistency is present (mixing geographic and temporal elements), reflecting limitations of the underlying model's knowledge rather than decoding instability.
This illustrates that token maturation primarily stabilizes \emph{generative dynamics}: by decoupling immediate token commitment from structural organization, the liquid tail enables even small models to sustain complex syntactic constructions.
This effect is consistent with the K-sweep results on Granite~3B (Sec.~\ref{sec:k_sweep}), where larger $K$ values sustain longer, more coherent outputs.

\subsection{Emergent Phrase-Level Commitment (Avalanche Locking)}
\label{app:avalanche}

A core property of Projected Autoregression is that the liquid tail allows the model to defer commitment until sufficient semantic context is available.
In our lock traces, we observe \emph{bursty} stabilization patterns in which multiple tail positions reach their final tokens within a narrow window of maturation stages, effectively behaving as a semantic ``chunk.''

Table~\ref{tab:avalanche} illustrates this phenomenon for the fragment ``\textit{'d arrive home around 6 AM}''.
Two recurring patterns emerge: (i)~phrase-level burst stabilization and (ii)~cascading stabilization downstream.

\paragraph{Burst stabilization.}
The tokens \texttt{'}, \texttt{d}, \texttt{arrive}, and \texttt{home} exhibit a descending lock-stage pattern ($4/4 \to 3/4 \to 2/4 \to 1/4$).
This is consistent with a single phrase-level disambiguation event: once a coherent interpretation of the verb phrase emerges, multiple positions rapidly converge toward their final tokens rather than drifting gradually.

\paragraph{Cascading stabilization.}
Early stabilization of an ``anchor'' token coincides with faster stabilization of later positions.
\texttt{home} stabilizes extremely early (stage $1/4$); following this, the numerical token \texttt{6} reaches its final form by stage $2/4$, and \texttt{AM} stabilizes at stage $2/4$ in alignment with \texttt{around}.
We interpret this as cascading stabilization: as earlier uncertainty collapses, the downstream context becomes more constrained, allowing later tokens to settle using fewer remaining stages.
While this does not establish strict causality, it supports the view that the liquid tail evolves as a coupled state---analogous to the centroid-convergence mechanism observed on Granite~3B (Sec.~\ref{sec:mechanistic}), where neighborhood narrowing at one position constrains neighboring positions.

\begin{table}[ht!]
    \centering
    \caption{\textbf{Avalanche Commitment and Cascading Stability (GPT-2).} Lock stages for ``\textit{'d arrive home around 6 AM}''. The early stabilization of \texttt{home} (stage $1/4$) is accompanied by earlier settling of downstream tokens.}
    \label{tab:avalanche}
    \vspace{0.2cm}
    \begin{small}
    \begin{tabular}{l c c l}
        \toprule
        \textbf{Token} & \textbf{Status} & \textbf{Lock Stage} & \textbf{Dynamics} \\
        \midrule
        I & Fixed & 4/4 & Prefix boundary \\
        \midrule
        \textbf{'} & \textbf{Fixed} & \textbf{4/4} & \multirow{3}{*}{\textit{Burst / Chunk}} \\
        \textbf{d} & \textbf{Fixed} & \textbf{3/4} & \\
        \textbf{arrive} & \textbf{Fixed} & \textbf{2/4} & \\
        \textbf{home} & \textbf{Fixed} & \textbf{1/4} & \textbf{$\leftarrow$ Early Anchor} \\
        \midrule
        around & Fixed & 4/4 & \\
        \textbf{6} & \textbf{Fixed} & \textbf{2/4} & \textbf{$\leftarrow$ Early settling after \texttt{home}} \\
        \textbf{AM} & \textbf{Fixed} & \textbf{2/4} & \textbf{$\leftarrow$ Early settling with \texttt{around}} \\
        \bottomrule
    \end{tabular}
    \end{small}
\end{table}

\subsection{Retroactive Disambiguation (Future-to-Past Stabilization)}
\label{app:retroactive}

We observe a suggestive pattern we term \textbf{retroactive stabilization}: an earlier token can remain liquid (unstable) until a later token in the tail provides disambiguating context.
The underlying transformer remains strictly causal; the effect arises because multiple tail positions are maintained and refined jointly before discrete commitment.
We present this as an exploratory observation.

Table~\ref{tab:reverse_lock} illustrates this for the fragment ``\texttt{º / 60 degrees Fahrenh}'', where the appearance of a strong unit token (\texttt{Fahrenh}) aligns with the stabilization of earlier symbols.

\paragraph{Reverse dependency analysis.}
In standard greedy decoding, tokens are finalized immediately in temporal order.
In Projected Autoregression, several positions remain undecided simultaneously, enabling later context to influence \emph{when} earlier positions stabilize:
\begin{itemize}[nosep]
    \item \textbf{Future anchor:} \texttt{Fahrenh} stabilizes early (stage $2/6$), indicating high semantic certainty for the unit.
    \item \textbf{Delayed symbol:} \texttt{º} (4 positions earlier) stabilizes only at stage $6/6$, suggesting its interpretation remained underdetermined until later context became available.
    \item \textbf{Stage--distance alignment:} The lock-stage gap ($6 - 2 = 4$) matches the positional distance, consistent with a same-step stabilization event.
\end{itemize}

The symbol \texttt{º} is semantically ambiguous in isolation (degree sign, ordinal indicator, geometric symbol).
The trace suggests the model keeps \texttt{º} liquid until \texttt{Fahrenh} appears in the tail, after which convergence becomes decisive.
The observed path efficiencies (67\%--73\%) indicate that while stabilization may be delayed, the eventual convergence is geometrically direct once disambiguating context appears.

This phenomenon is consistent with the broader framework: the liquid tail functions as a coupled dynamical system where positions influence each other through attention, enabling non-local disambiguation that is impossible under immediate commitment.

\begin{table}[ht!]
    \centering
    \caption{\textbf{Retroactive Locking Trace (GPT-2).} A ``reverse cascade'' pattern where stabilization appears to propagate from a later unit token (\texttt{Fahrenh}) to earlier symbols (\texttt{º}).}
    \label{tab:reverse_lock}
    \vspace{0.2cm}
    \begin{small}
    \begin{tabular}{l c c l}
        \toprule
        \textbf{Token} & \textbf{Lock Stage} & \textbf{Path Eff.} & \textbf{Dependency} \\
        \midrule
        \textbf{Fahrenh} & \textbf{2/6} & \textbf{70\%} & \textbf{Future Anchor (Trigger)} \\
        degrees & 4/6 & 66\% & $\downarrow$ Stabilizes with \texttt{/} \\
        60 & 6/6 & 68\% & \\
        / & 6/6 & 73\% & $\uparrow$ Stabilizes with \texttt{degrees} \\
        \textbf{º} & \textbf{6/6} & \textbf{67\%} & \textbf{$\uparrow$ Stabilizes with \texttt{Fahrenh}} \\
        \bottomrule
    \end{tabular}
    \end{small}
\end{table}

% ============================================================
% H. DISTRIBUTIONAL COMPARISON
% ============================================================

\section{MAUVE with OpenWebText Reference}
\label{sec:app_openwebtext}

MAUVE~\cite{pillutla2021mauve} measures the distributional similarity between generated text and a reference corpus.
A critical but often overlooked factor is the choice of reference corpus: MAUVE scores are meaningful only when the reference represents the distribution the model was trained on.
Our models are trained on FineWeb, so FineWeb is the natural reference.
However, many prior works report MAUVE against OpenWebText, so for completeness we report scores against both.

The results reveal substantial sensitivity to this choice.
Against FineWeb (the shared training distribution), Projected Autoregression is competitive with AR baselines.
Against OpenWebText, the ranking shifts dramatically: AR methods score much higher because vanilla Granite~3B (without the AR LoRA adapter) was pretrained on data closer to the OpenWebText distribution, giving it a natural advantage on this reference regardless of generation quality.

\begin{table}[ht!]
\centering
\small
\begin{tabular}{lcc}
\toprule
\textbf{Method} & \textbf{MAUVE (FineWeb)} & \textbf{MAUVE (OWT)} \\
\midrule
Proj.\ AR & 0.849 & 0.452 \\
Greedy$^\dagger$ & 0.749 & 0.714 \\
Greedy + rep$^\dagger$ & 0.881 & 0.964 \\
Top-$p$=0.9$^\dagger$ & 0.869 & 0.993 \\
Top-$p$=0.95$^\dagger$ & 0.832 & 0.921 \\
\bottomrule
\end{tabular}
\caption{MAUVE under different reference corpora.
$^\dagger$OpenWebText column uses vanilla Granite~3B (not AR~LoRA).
Against FineWeb (shared training distribution), Proj.\ AR is competitive; against OpenWebText, the ranking shifts.}
\label{tab:mauve_owt}
\end{table}

The key takeaway is that MAUVE scores should be interpreted relative to the training distribution.
Projected Autoregression achieves MAUVE~$= 0.849$ against FineWeb, which is competitive with AR baselines (0.749--0.881 excluding top-$p$).
The lower score against OpenWebText (0.452) does not indicate poor generation quality---it reflects a distributional mismatch between FineWeb-trained text and the OpenWebText reference.
We recommend that future work on continuous-space generation report MAUVE against the actual training distribution to avoid misleading comparisons.

\end{document}